\documentclass[letterpaper, 10 pt, conference]{ieeeconf}

\IEEEoverridecommandlockouts
\usepackage{amsmath}
\usepackage{amssymb}
\usepackage[acronym]{glossaries}
\usepackage{graphicx}
\usepackage{xcolor}
\usepackage{booktabs}
\usepackage{pifont}
\usepackage{algorithm}
\usepackage{stfloats}
\usepackage{cuted}
\usepackage{ragged2e}
\usepackage{svg}
\usepackage{subcaption}
\usepackage{threeparttable}
\usepackage{hyperref}

\usepackage{listings}

\newcommand{\cmark}{\ding{51}}

\newacronym{rl}{RL}{Reinforcement Learning}
\newacronym{marl}{MARL}{Multi-Agent Reinforcement Learning}
\newacronym{harl}{HARL}{Heterogeneous Agent Reinforcement Learning}
\newacronym{gpus}{GPUs}{graphics processing units}
\newacronym{cpus}{CPUs}{central processing units}
\newacronym{ppo}{PPO}{Proximal Policy Optimization}
\newacronym{happo}{HAPPO}{Heterogeneous Agent Proximal Policy Optimization}
\newacronym{mlp}{MLP}{multi-layer perceptron}
\newacronym{maddpg}{MADDPG}{multi-agent deep deterministic policy gradient}
\newacronym{ctde}{CTDE}{Centralized Training and Decentralized Execution}
\newacronym{api}{API}{application programming interface}
\newacronym{mdp}{MDP}{Markov Decision Process}

\title{\LARGE \bf
HARL-A: An Extensible Benchmark Framework for Heterogeneous Multi-Agent Adversarial Reinforcement Learning in IsaacLab
}

\author{
  Isaac Peterson\textsuperscript{*1} \and
  Christopher Allred\textsuperscript{*12}\thanks{* Equal Contributions.}\thanks{1 Peterson, Allred, Morrey, and Harper are affiliated with Utah State University.}\thanks{2 Allred is affiliated with US DEVCOM Army Research Laboratory.} \and
  Jacob Morrey\textsuperscript{1} \and
  Mario Harper\textsuperscript{1}
}

\begin{document}
\maketitle


\begin{abstract}
Progress in adversarial multi-agent reinforcement learning (\gls{marl}) for robotics has been hampered by a lack of shared, extensible infrastructure that supports heterogeneous agent morphologies in high-fidelity physics simulation. Existing frameworks either focus on cooperative tasks, rely on simplified physics engines, or provide isolated implementations that are difficult to extend. We present HARL-A, an open-source, actively maintained framework built on IsaacLab that enables scalable training and benchmarking of adversarial policies across morphologically diverse robot teams with any number of teams and any mix of robot morphologies per team. HARL-A extends the HARL algorithm library and IsaacLab with adversarial multi-agent support and contributes three components: (1) a modular software architecture that reduces the engineering overhead of defining new heterogeneous adversarial environments, (2) a suite of three benchmark environments---Sumo, Soccer, and 3D Galaga---spanning contact-rich pushing, ball-skill competition, and pursuit/evasion, (3) over ten pretrained policies spanning homogeneous and heterogeneous team configurations, released publicly on Hugging Face to enable immediate exploration of adversarial learning dynamics without retraining from scratch. We demonstrate the framework across multiple competitive scenarios, showing that it reliably produces learned adversarial policies and emergent role specialization. All code environments, trained policies, and documentation are openly available at https://github.com/DIRECTLab/IsaacLab-HARL.
\end{abstract}


\section{Introduction}

The development of robust multi-agent policies for competitive robotics applications---pursuit-evasion, adversarial manipulation, security---requires training infrastructure that combines three properties rarely found together: high-fidelity contact physics, support for heterogeneous robot morphologies, and scalable parallelized simulation that can train adversarially. Without the first, learned policies fail to transfer to real mechanical systems. Without the second, benchmarks cannot capture the strategic asymmetries that make real-world competition interesting. Without the third, the sample complexity of adversarial training makes meaningful experiments impractical.

\begin{figure}
\centering
\includegraphics[width=1\linewidth]{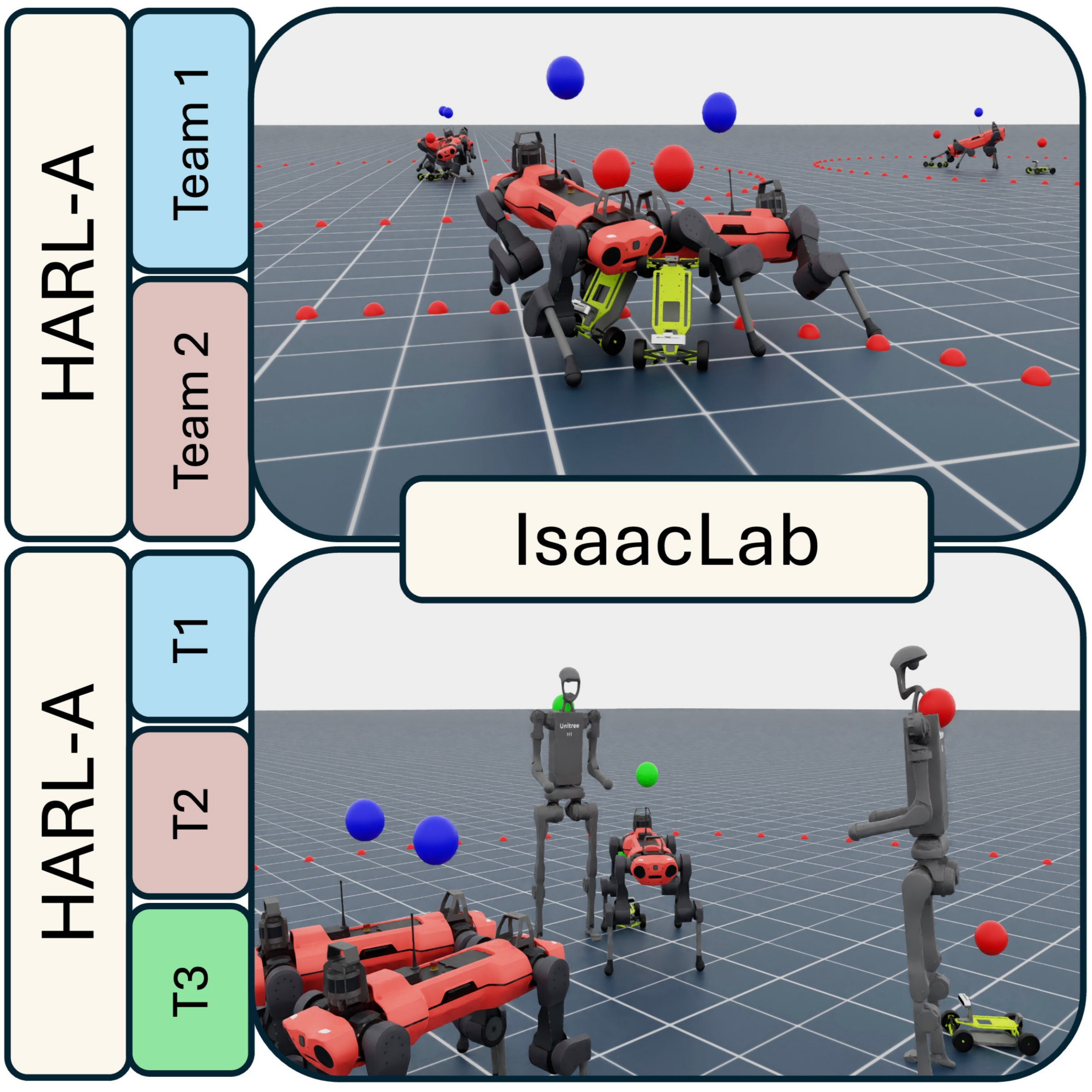}
\caption{HARL-A environments in IsaacLab. (Top) Two heterogeneous teams---each comprising one Anymal C quadruped and one Leatherback rover---compete in an adversarial Sumo task. (Bottom) Multi-team configurations with three distinct agent morphologies illustrate the framework's support for arbitrary team composition and scale.}
\label{fig:multiple_environments}
\end{figure}

Reinforcement learning has matured rapidly for single-robot tasks \cite{kober2013reinforcementlearninginroboticsurvey, han2023deeprlforroboticssurvey}, producing capable locomotion \cite{ilija2024humanoid, choi2023quadrapedallocamotion} and dexterous manipulation policies \cite{Chen2022}. GPU-accelerated simulators such as IsaacLab \cite{mittal2023orbit} have made large-scale training feasible, and recent work has extended this to cooperative heterogeneous \gls{marl} \cite{haight2025heterogeneous}. However, adversarial settings have received comparatively little attention in high-fidelity simulation. Prior adversarial work has largely relied on simplified physics \cite{Bansal2018, Baker2020} or has addressed only homogeneous agents \cite{Li2025}. No existing framework supports extensible, heterogeneous, adversarial training in a physics-accurate simulator.

We address this gap with HARL-A (Heterogeneous Agent Reinforcement Learning -- Adversarial), a framework that extends IsaacLab and the HARL library \cite{zhong2024HARL} to support competitive multi-team training. The key design insight is that adversarial heterogeneous settings require two architectural changes that cooperative frameworks do not: team-isolated critic networks and a team-indexed environment \gls{api} that cleanly separates observations, actions, and rewards by team. HARL-A implements both, and wraps them in a modular environment interface that allows new scenarios to be defined with minimal boilerplate.

The primary contributions of this work are:
\begin{enumerate}
    \item \textbf{An adversarial extension of the HARL algorithm library}, adding team-isolated critics and a team-indexed environment \gls{api} that supports any number of competing teams with any mix of robot morphologies. Other \gls{marl} algorithms can be integrated and tested against the same environments, and all code is open-source and actively maintained.
    \item \textbf{A benchmark suite} of three adversarial environments--Sumo, Soccer, and 3D Galaga--covering distinct task categories with both homogeneous and heterogeneous team variants. The benchmark is designed to expose qualitatively different competitive challenges, from contact-rich pushing to long-horizon pursuit, enabling systematic evaluation of adversarial \gls{marl} algorithms.
    \item \textbf{Over ten pretrained policies} released on Hugging Face, spanning curriculum stages, team compositions, and morphology combinations (Table~\ref{tab:policies}). These enable researchers to immediately explore adversarial learning dynamics (visualizing emergent behaviors, running cross-play evaluations, or fine-tuning for new scenarios) without retraining from scratch.
\end{enumerate}

The framework is designed to serve robotics researchers who want to add new robot morphologies or competitive tasks to the benchmark, and \gls{marl} algorithm developers who need physically grounded, heterogeneous adversarial environments for evaluating new methods. In both cases, the combination of a flexible \gls{api}, GPU-scale throughput, and a library of pretrained starting points substantially lowers the barrier to entry.


\section{Related Work}
\label{sec:related_works}

\begin{table*}[t]
    \centering
    \footnotesize
    \begin{threeparttable}
    \begin{tabular}{|c|c|c|c|c|c|c|}
        \hline
        \textbf{Author (Year)} & \textbf{Environment} & \textbf{Adversarial} & \textbf{High-Fidelity Physics} & \textbf{Multi-Agent} & \textbf{Heterogeneous} & \textbf{Extensible Framework} \\
        \hline
        Bansal et al. (2018) \cite{Bansal2018} & MuJoCo & \cmark & & & & \\
        \hline
        Gleave et al. (2020) \cite{Gleave2020Adversarial} & MuJoCo & \cmark & & & & \\
        \hline
        Lowe et al. (2017) \cite{lowe2020multiagentactorcriticparticle} & Particle Env. & \cmark & & \cmark & & \\
        \hline
        Baker et al. (2020) \cite{Baker2020} & MuJoCo & \cmark & & \cmark & & \\
        \hline
        Schwarting et al. (2021) \cite{Schwarting2021DeepLC} & Visual RL & \cmark & & \cmark & & \\
        \hline
        Liu et al. (2021) \cite{liu2021} & MuJoCo & \cmark & & \cmark & & \\
        \hline
        Li et al. (2025) \cite{Li2025} & IsaacLab & \cmark & \cmark & \cmark & & \\
        \hline
        \textbf{HARL-A (Ours)} & IsaacLab & \cmark & \cmark & \cmark & \cmark & \cmark \\
        \hline
    \end{tabular}
    \end{threeparttable}
    \caption{Comparison of adversarial MARL works. HARL-A is the only framework that simultaneously satisfies all five properties.}
    \label{tab:related_work_comparison}
\end{table*}

\subsection{Adversarial Multi-Agent Learning}

Early work demonstrated that self-play in physics-based environments could produce emergent competitive behaviors \cite{Bansal2018}, and subsequent work showed that adversarial training could exploit brittle policies that are undetectable under standard evaluation \cite{Gleave2020Adversarial}. Multi-agent extensions introduced \gls{maddpg} \cite{Lowe2017} and showed that large-scale population-based training could reach human-level performance in complex games \cite{Berner2019Dota2W, Vinyals2019rlforstarcraft}. The MuJoCo hide-and-seek environment \cite{Baker2020} demonstrated rich emergent tool use from repeated self-play. More recent work combined imitation learning, \gls{marl}, and population training to produce competitive humanoid soccer \cite{liu2021}.

\subsection{Benchmarks and Frameworks}

The \gls{marl} community has produced several benchmark frameworks, including VMAS \cite{}, SMACv2 \cite{}, and Multi-Agent Particle Environments \cite{lowe2020multiagentactorcriticparticle}. However, these rely on simplified or 2D physics and do not support morphologically heterogeneous agents. In the high-fidelity space, MARLaDona \cite{Li2025} demonstrated competitive robot soccer in IsaacLab, but focuses on homogeneous teams and does not provide an extensible framework for new environments or morphologies. The cooperative counterpart to our work \cite{haight2025heterogeneous} introduced heterogeneous multi-agent training in IsaacLab, but was limited to cooperative tasks with a shared critic. HARL-A extends this line of work to adversarial settings and provides the benchmark infrastructure the community currently lacks, as summarized in Table~\ref{tab:related_work_comparison}.

\subsection{Curriculum Learning for Complex Tasks}

Training stability in competitive settings is a well-recognized challenge \cite{Bansal2018, narvekar2020cirriculumlearning}. Procedural content generation methods, such as PAIRED \cite{dennis2020advesrarialenvPAIRED}, adaptively modulate task difficulty, while techniques like domain randomization and progressive task decomposition \cite{narvekar2020cirriculumlearning} have demonstrated strong performance in locomotion tasks. In this work, we extend the zero-buffer curriculum framework proposed in \cite{wang2020few}, which preserves a consistent observation space across curriculum stages and facilitates smooth policy transfer between them.


\section{Framework Design}
\label{sec:framework}

\begin{figure*}[t]
    \centering
    \includegraphics[width=.99\linewidth]{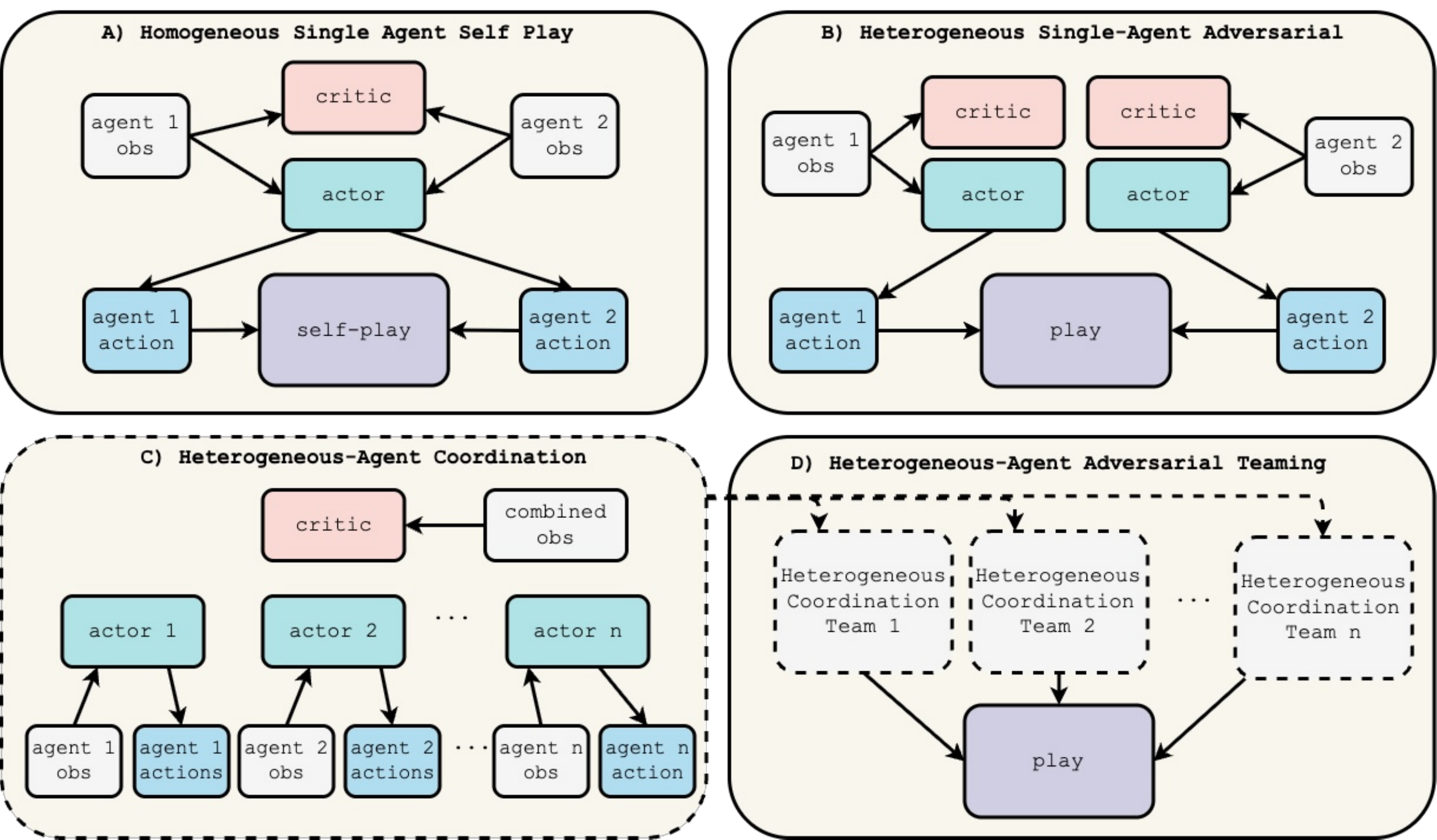}
    \caption{Actor-critic training paradigms in adversarial reinforcement learning. Prior cooperative frameworks supported paradigm C (heterogeneous coordination with shared critic). HARL-A adds support for paradigms B (heterogeneous single-agent adversarial) and D (heterogeneous multi-team adversarial), enabling the full spectrum of competitive training configurations.}
    \label{fig:trainingadversarial}
\end{figure*}

HARL-A is built on two existing systems: IsaacLab \cite{mittal2023orbit}, which provides GPU-accelerated physics simulation, and the HARL library \cite{zhong2024HARL}, which provides heterogeneous actor-critic algorithms. Our contribution is the adversarial extension connecting them, comprising three design components described below.

\subsection{Team-Indexed Environment API}
HARL-A exposes a simple dictionary in the IsaacLab configuration class that allows for the configuration of the teams for adversarial training. 
\begin{lstlisting}
  teams = {
      "team_0": ["anymal_0", "anymal_1", ],
      "team_1": ["leatherback_0", "leatherback_1"
      ...
      "team_N": ["robot_A_N", ... "robot_N_N"
  }
\end{lstlisting}


This central interface change in HARL-A is a shift from agent-indexed to team-indexed environment interaction. In cooperative \gls{marl} frameworks, observations, actions, and rewards are typically organized as flat per-agent arrays. In adversarial settings, this is insufficient: teams require isolated reward signals and critics, and the environment must expose which agents belong to which team to support \gls{ctde}.

This modular abstraction enforces a clear separation between task specification and learning infrastructure, reducing coupling between environment logic and optimization components. As a result, modifications to morphology or task dynamics remain localized to a small portion of the codebase, lowering the risk of unintended side effects in the training pipeline. By reusing the same parallelization backend, logging utilities, and curriculum mechanisms across tasks, the framework minimizes duplicated implementation effort and improves experimental reproducibility. In practice, this substantially reduces development overhead and iteration time, enabling rapid prototyping of new competitive scenarios while maintaining a consistent evaluation and training protocol.

\subsection{Team-Specific Critics}

A cooperative framework's shared critic is insufficient for adversarial settings. When team rewards are zero-sum (i.e., $r^{(0)}_t = -r^{(1)}_t$), a single critic estimating $V(s_t)$ collapses toward zero regardless of which team is winning, producing vanishing advantage estimates and no useful gradient signal. HARL-A addresses this by assigning each team its own critic $V^{(i)}(s)$ trained against its own reward $r^{(i)}$, consistent with the HAPPO formulation \cite{zhong2024HARL}. This is the minimal architectural change needed to make standard actor-critic methods functional in zero-sum settings, and it composes cleanly with the team-indexed API described above.

\subsection{Training Regime Support}

HARL-A supports three adversarial training regimes out of the box, selectable via configuration:

\begin{itemize}
    \item \textbf{Alternating (``leapfrog'')}: One team's actor is frozen while the other updates for a fixed number of timesteps, then roles switch. This mirrors the regime used in \cite{lerrel2017adversarialreinforcementlearning} and generally produces stable learning curves.
    \item \textbf{Simultaneous}: Both teams update concurrently. While less stable theoretically, we find that it also produces meaningful policies in practice, suggesting that the framework is robust to this choice.
    \item \textbf{Fixed-opponent}: One team is held at a fixed checkpoint while the other trains. Useful for measuring improvement against a defined baseline.
\end{itemize}

All three regimes use the same environment and critic infrastructure, making it straightforward to compare them for a given task.


\section{Benchmark Environments}
\label{sec:environments}


HARL-A ships with three benchmark environments spanning distinct task categories. Each environment provides homogeneous and heterogeneous team variants, a multi-stage curriculum, and standardized evaluation metrics (win rate against a fixed baseline policy). The primary robot platforms are the Anymal C quadruped and the Leatherback wheeled rover, selected to maximize morphological contrast: Anymal C features high-dimensional joint control and contact-rich locomotion dynamics, whereas Leatherback uses simple differential-drive control with high translational speed, naturally inducing differentiated strategic roles during adversarial training. To demonstrate generality beyond these canonical platforms, we additionally incorporate the Minitank, a custom morphology developed from scratch, and a quadrotor-style drone. These agents integrate without modification to the training pipeline and are showcased in the 3D Galaga benchmark, a pursuit-and-interception scenario in which competitive asymmetry is defined directly by task structure rather than embodiment alone, further illustrating HARL-A’s ability to compose heterogeneous robot skills within a unified framework.

\subsection{Sumo}

\begin{figure}[t]
    \centering
    \includegraphics[width=1.0\linewidth]{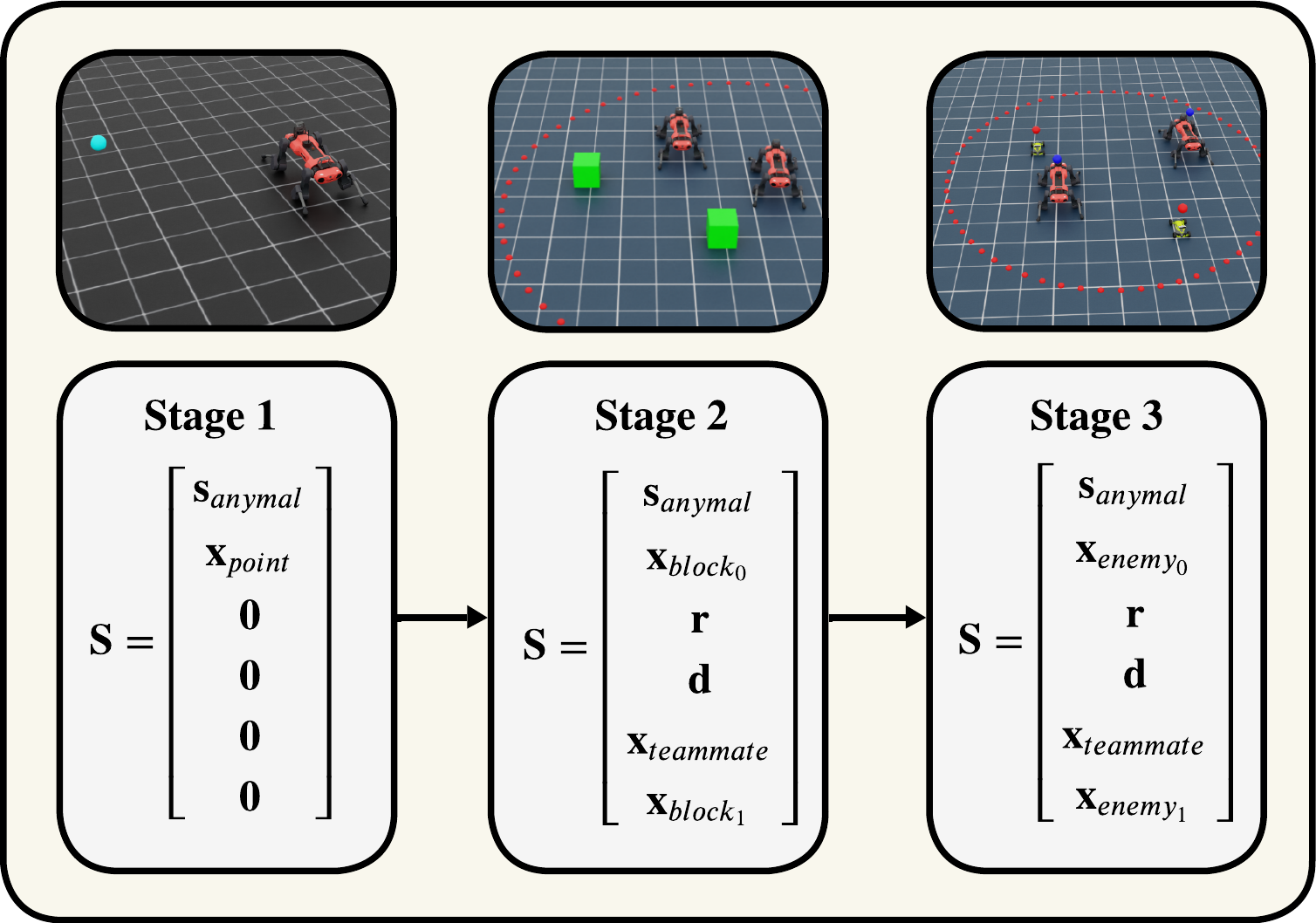}
    \caption{Evolution of the observation space across Sumo curriculum stages for the Anymal C robot. Zero-padded entries (shown as \textbf{0}) are replaced with meaningful features as task complexity increases, following the zero-buffer strategy of \cite{wang2020few}. This maintains a fixed observation dimension across all stages, enabling seamless policy transfer. This strategy was used for all benchmark environments presented in this paper.}
    \label{fig:advs_tasks}
\end{figure}

The Sumo environment is the primary benchmark for contact-rich adversarial interaction. Two teams compete to push all opposing robots out of a circular arena. The environment decomposes into three curriculum stages.

\noindent\textbf{Stage 1 --- Locomotion.} Before competing, the Anymal C agent must first acquire stable locomotion. The reward combines walking shaping terms $W$ with a goal-reaching term:
\[
R(S_t) = \sum_{i=0}^{|W|}W_i + \delta\,\mathbf{1}_{\text{reached\_goal}} + \gamma\bigl(1 - \tanh(\alpha\,\mathbf{d})\bigr)
\]
where $\mathbf{d}$ is distance to goal and $\delta, \gamma, \alpha$ are reward scalars.

\noindent\textbf{Stage 2 --- Object Pushing.} Each robot learns to push a block out of the arena in isolation, before the opponent is introduced. The reward is:
\[
R(S_t) = (\hat{r}_t + \hat{d}_t)\,\Delta t + T + \delta\bigl(\mathbf{1}_{\text{push\_out}} - \mathbf{1}_{\text{left\_ring}}\bigr)
\]
where $\hat{r}_t = \tanh(r_t / r_{\max})$ rewards moving the block toward the boundary, $\hat{d}_t = 1 - \tanh(d_t / r_{\max})$ rewards approaching the block, and $T$ is a step penalty for efficiency.

\noindent\textbf{Stage 3 --- Adversarial Sumo.} The block is removed and teams compete directly. Each team $i$ receives:
\begin{equation}
R_i(S_t) = \tau \cdot \bigl(L_j - L_i - \phi\bigr) \cdot \kappa
\label{eq:sumo_reward}
\end{equation}
where $L_i = 1$ if any agent on team $i$ has left the ring, $\tau = 0$ on simultaneous elimination (tie), $\phi = 1$ on timeout, and $\kappa$ is a reward scaling constant. This formulation penalizes self-elimination, rewards opponent elimination, and prevents reward on draws.

The zero-buffer strategy (Figure~\ref{fig:advs_tasks}) maintains a fixed-dimension observation vector across all stages by padding unused entries with zeros. This allows a policy trained at Stage~1 to be directly hot-started at Stage~2 without architectural changes.

\subsection{Soccer}

\begin{figure}[t]
    \centering
    \includegraphics[width=1.0\linewidth]{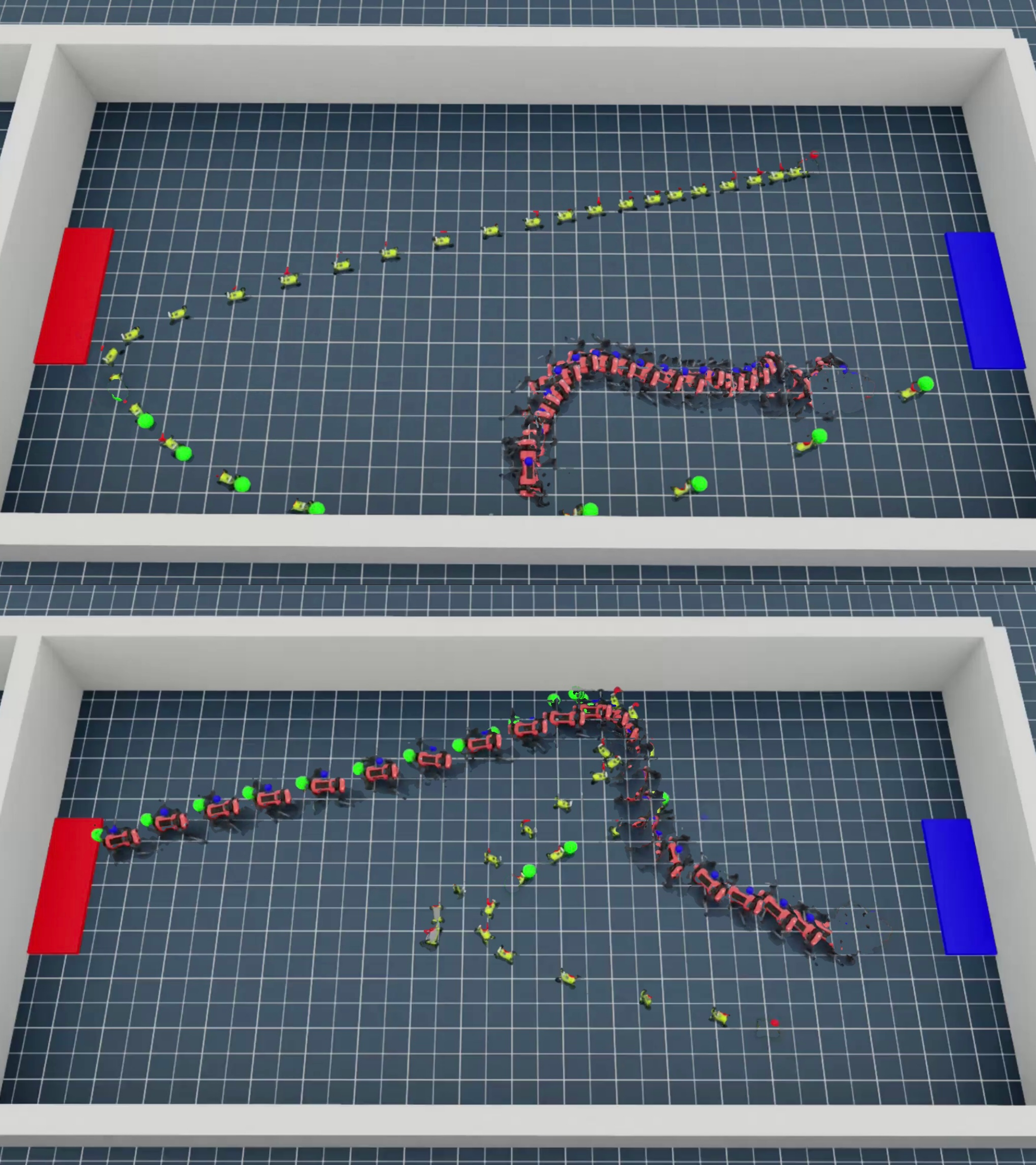}
    \caption{Competitive behaviors in the heterogeneous Soccer environment (Anymal C vs.\ Leatherback). Top: Leatherback maintains possession and scores. Bottom: Leatherback possession followed by a turnover and an Anymal goal, illustrating the competitive dynamics that emerge from morphological asymmetry.}
    \label{fig:soccer_goals}
\end{figure}

The Soccer environment benchmarks ball-skill competition and longer-horizon strategy. Teams score by driving a ball through the opponent's goal plane. The observation space includes robot--ball and ball--goal vectors, agent velocities, teammate and opponent poses, and arena context. Actions are morphology-appropriate: joint velocity targets for Anymal C, wheel velocity commands for Leatherback.

The environment uses a two-stage curriculum. Stage~1 trains single-robot ball approach and driving skills using dense distance rewards and sparse goal events. Stage~2 introduces the opposing team with scoring rewards and losing penalties. Episode termination occurs on a goal or a configurable timeout.

We provide a trained 1-v-1 heterogeneous variant (one Anymal C vs.\ one Leatherback), trained with the alternating leapfrog regime. Figure~\ref{fig:soccer_goals} illustrates competitive play. Soccer complements Sumo by rewarding sustained ball control rather than contact-based elimination, testing a qualitatively different set of adversarial skills.

\subsection{3D Galaga}

\begin{figure}[t]
    \centering
    \includegraphics[width=1\linewidth]{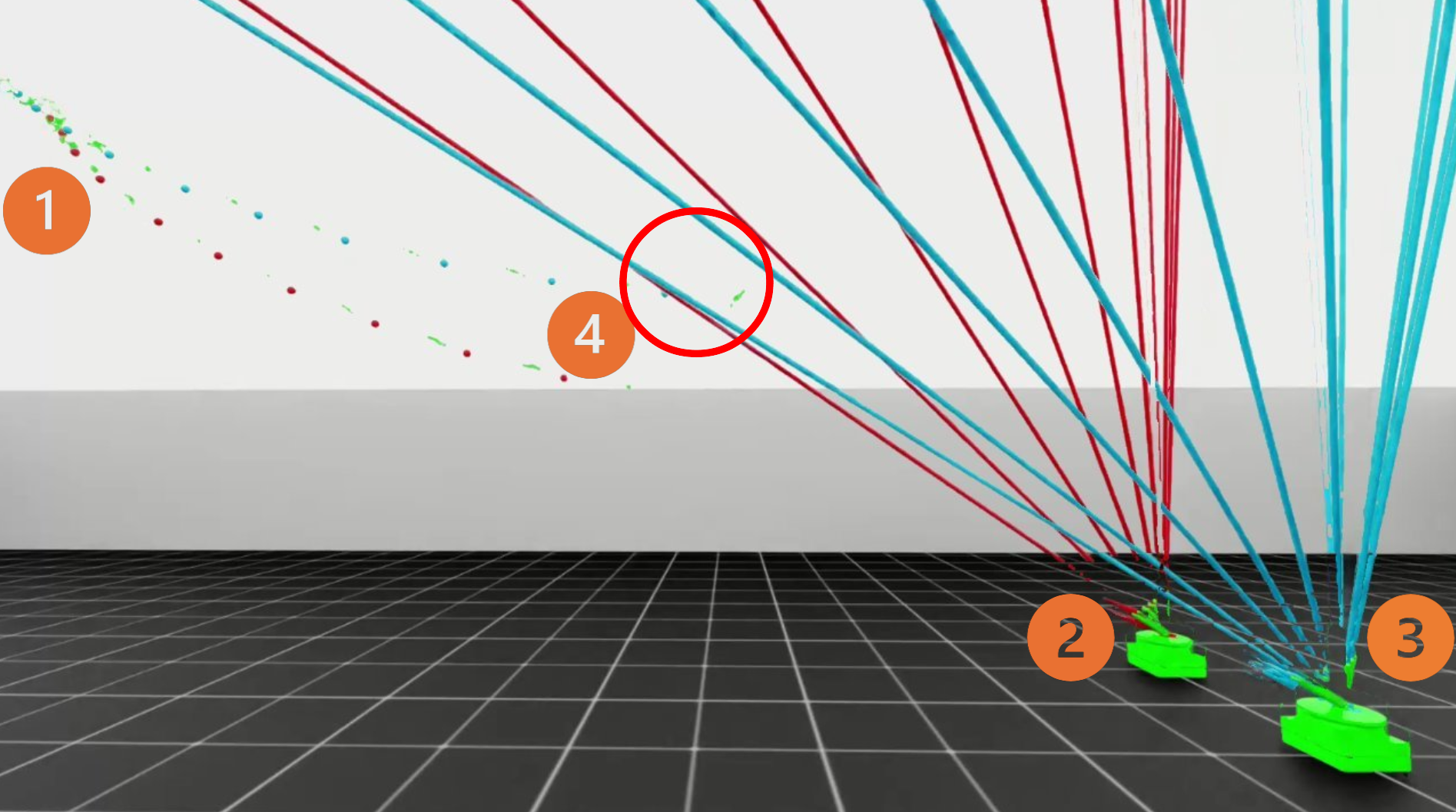}
    \caption{3D Galaga: Anti-Aircraft Defense. MiniTanks fire arm-aligned laser-tag rays (2, 3) each control step. A drone (path shown in 1) is knocked out when it enters the radius of an active ray (4). The environment demonstrates HARL-A's ability to compose pretrained locomotion primitives into novel adversarial scenarios.}
    \label{fig:mini3DG}
\end{figure}

The 3D Galaga environment demonstrates a third task category: pursuit and interception in three-dimensional space. A team of MiniTanks attempts to intercept a drone navigating to waypoints using arm-aligned laser-tag rays. The drone receives a reward for reaching within 0.2 meters of its target waypoint; each tank receives a reward for pinning the drone within a 0.02-radian cone off its arm.

In this environment the drone enters adversarial training with an existing waypoint-navigation policy, and the tanks are initialized with pointing policies rather than being trained from scratch. This illustrates one of HARL-A's design goals---supporting the composition of existing robot skills into new adversarial tasks with minimal engineering overhead. It also represents a qualitatively different competitive structure from Sumo and Soccer, as the asymmetry between offense (evasion) and defense (interception) is built into the task definition rather than emerging from morphological differences.


\section{Experiments}
\label{sec:results}

All experiments were run on  NVIDIA RTX PRO 6000 Blackwell series GPUs. We evaluate the framework along three axes: policy-learning quality (win rate), scalability (simulation throughput), and emergent-behavior diversity.

\subsection{Policy Learning}

\begin{figure}[!htb]
    \centering
    \begin{subfigure}[!htbp]{0.45\textwidth}
        \centering
        \includegraphics[width=\textwidth]{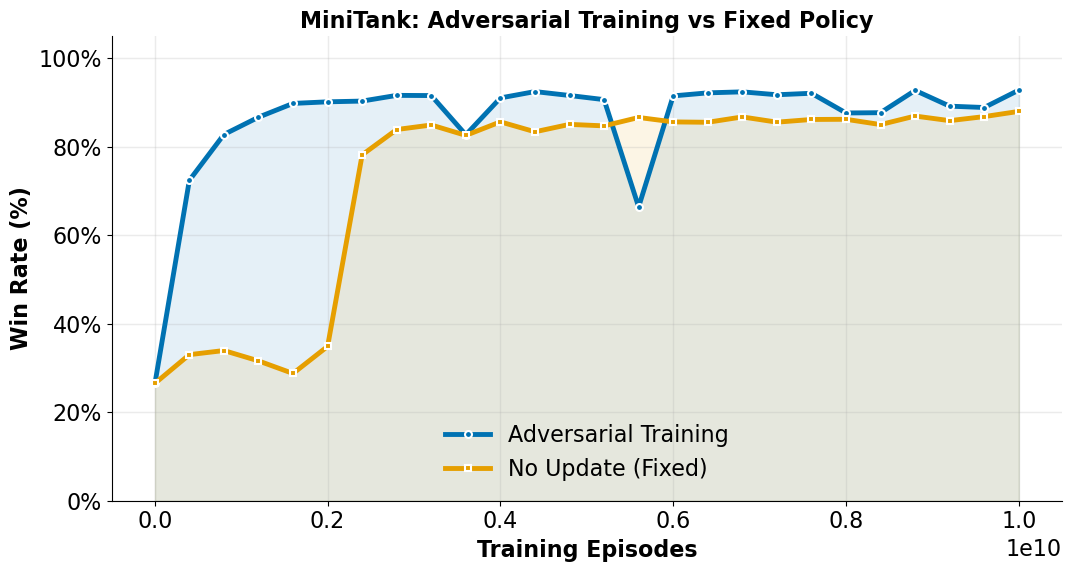}
        \label{fig:results_galaga}
    \end{subfigure}
    \hfill
    \begin{subfigure}[!htbp]{0.45\textwidth}
        \centering
        \includegraphics[width=\textwidth]{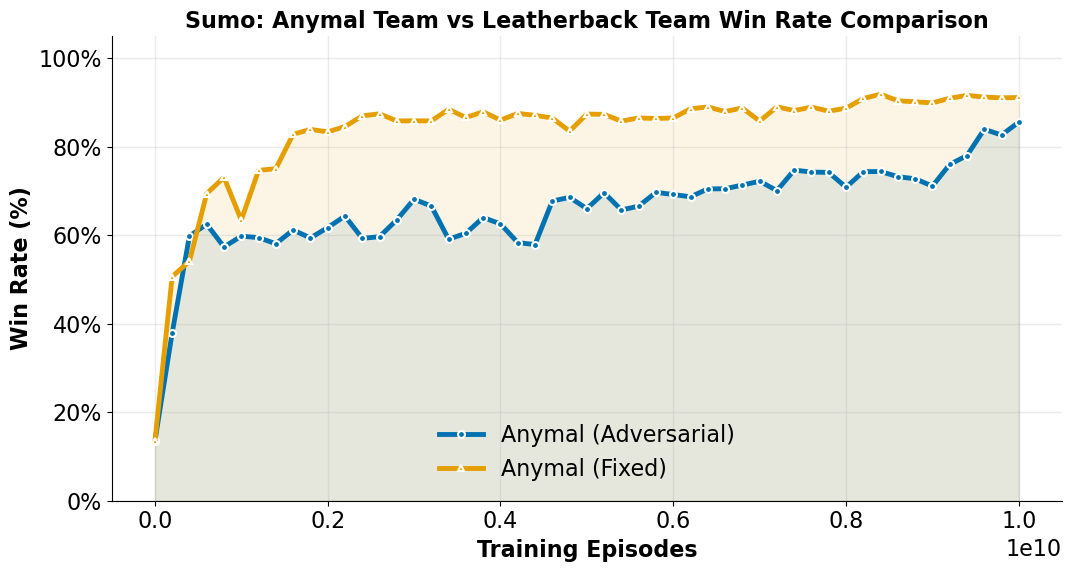}
        \label{fig:results_sumo}
    \end{subfigure}
    \begin{subfigure}[!htbp]{0.45\textwidth}
        \centering
        \includegraphics[width=\textwidth]{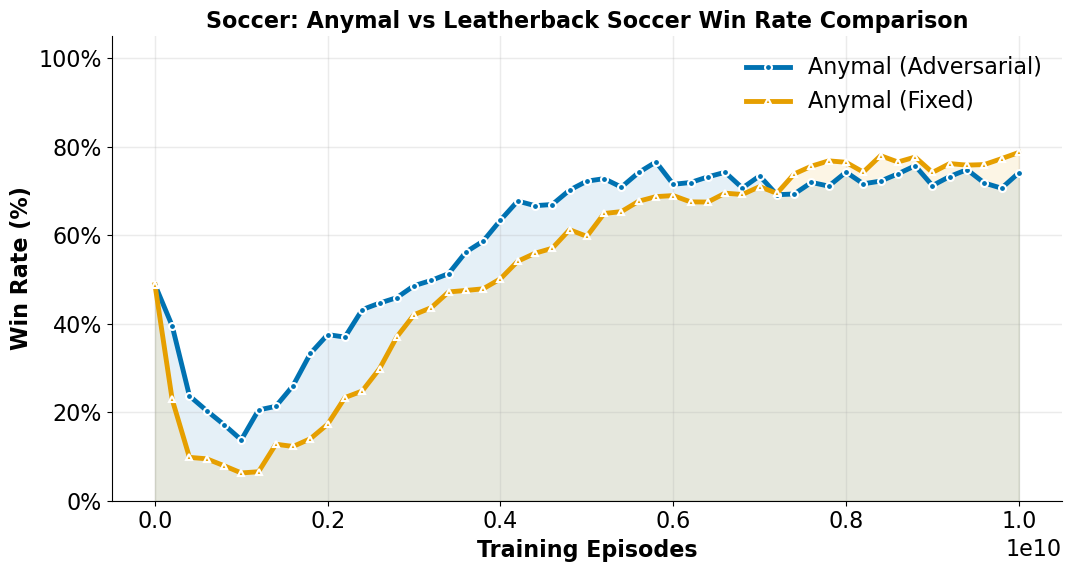}
        \label{fig:results_soccer}
    \end{subfigure}
    \caption{Win rates of adversarially trained policies vs.\ a fixed baseline policy across the three benchmark environments (top: 3D Galaga, middle: Sumo, bottom: Soccer). Blue shows the agents trained adversarially, while orange represents agents trained against a fixed policy. For each saved checkpoint, we evaluate 1000 parallel environments against the initial baseline policy and average the win rate. Increasing win rate over training confirms that meaningful adversarial learning occurs across all three environments and task types.}
    \label{fig:main_figure}
\end{figure}

Figure~\ref{fig:main_figure} shows win rate curves across all three benchmark environments. These curves demonstrate the effectiveness of the training, as in both cases, whether trained in an adversarial environment or against a fixed policy, the agent learned to exploit weaknesses in the baseline policy. Following standard practice in adversarial training evaluation \cite{Berner2019Dota2W, Bansal2018}, we report win rate against a fixed initial policy rather than against a co-evolving opponent, which provides a cleaner measure of absolute policy improvement.

\subsection{Emergent Adversarial Behaviors}

\begin{figure*}[t]
    \centering
    \includegraphics[width=1\linewidth]{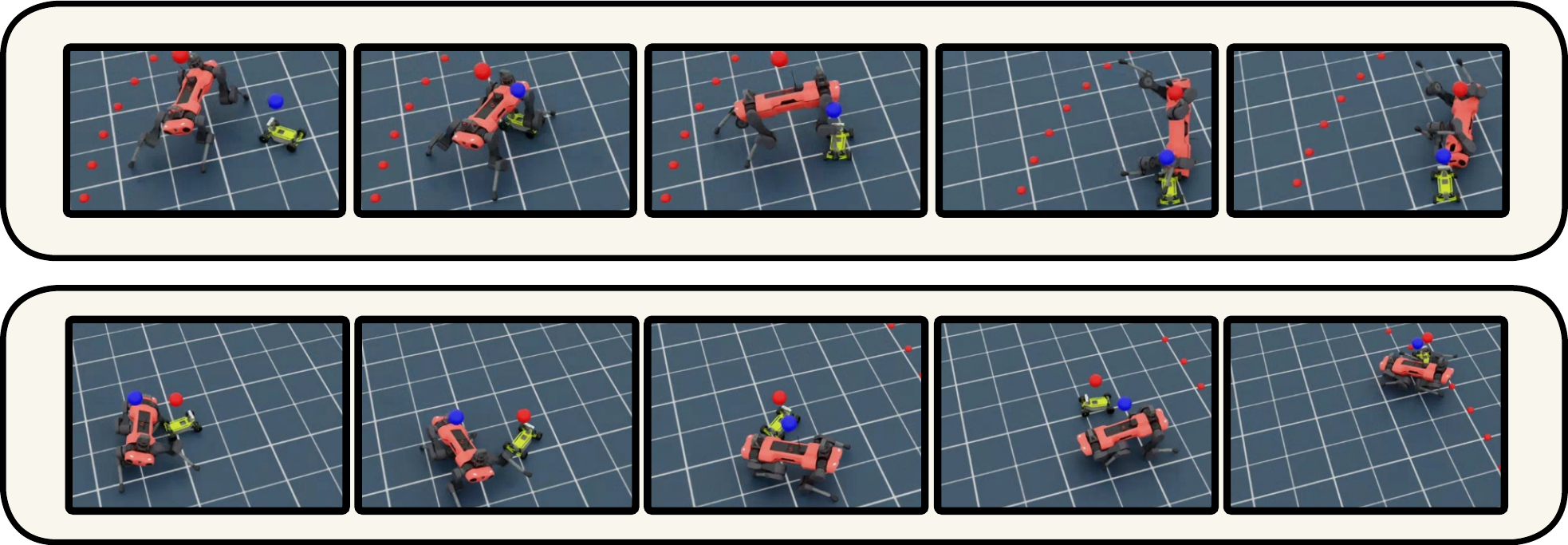}
    \caption{Emergent role specialization in the heterogeneous Sumo environment (one Anymal C and one Leatherback per team). Without any explicit role assignment, the Leatherback learned to destabilize the opposing Anymal by targeting its legs (Top), while the Anymal learned to drag the opposing Leatherback out of the arena using its limbs (Bottom). These complementary strategies emerged consistently from morphological asymmetry.}
    \label{fig:emergent_behavior}
\end{figure*}

Beyond win rates, training produced qualitatively distinct emergent behaviors across morphologies. In the heterogeneous Sumo environment, Leatherback rovers learned to target the legs of opposing Anymal robots---exploiting their low ground clearance---while Anymal robots developed dragging maneuvers that used their limbs to pull rovers outside the arena boundary (Figure~\ref{fig:emergent_behavior}). Critically, these complementary roles emerged without any explicit role assignment mechanism: the only signal was the team reward in Equation~\ref{eq:sumo_reward}. This suggests that morphological asymmetry alone is sufficient to induce division of labor in adversarial training, a finding of independent interest for multi-agent learning research.

These behaviors parallel emergent phenomena observed in large-scale self-play systems \cite{Baker2020}, but arise here in a high-fidelity contact-rich setting where physical constraints actively shape which strategies are viable.

\subsection{Scalability}

\begin{figure}[H]
    \centering
    \includegraphics[width=1\linewidth]{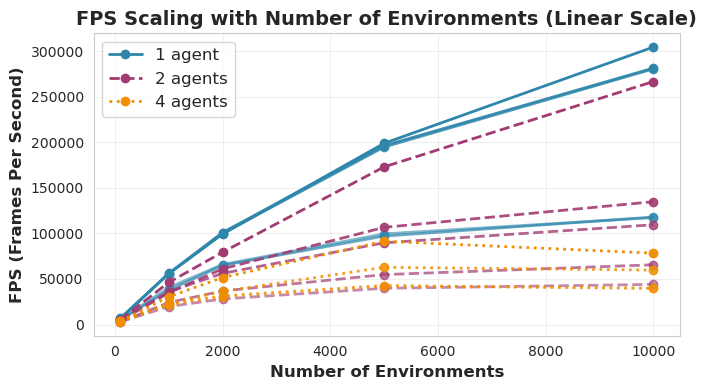}
    \caption{FPS training distribution based on the number of agents in each environment. As can be seen in the figure, increased agent count decreases the FPS of the training.}
    \label{fig:fps}
\end{figure}


Figure~\ref{fig:fps} shows simulation throughput across environments. IsaacLab's GPU parallelization yields substantial throughput gains that scale nearly linearly with the number of environments, up to a saturation point determined by the number of robots spawned per environment. For the four-robot heterogeneous Sumo configuration, throughput plateaus around 5000 parallel environments. Single-robot environments (e.g., Sumo Stage~1) continue scaling beyond 10000 environments. This relationship provides a useful rule of thumb for practitioners: throughput saturation occurs approximately when the total robot count (environments $\times$ robots per environment) exceeds the GPU's physics simulation budget.

These throughput numbers are practically significant for adversarial training: the win-rate curves in Figure~\ref{fig:main_figure} were obtained at $10^{10}$ training steps, which would be computationally prohibitive without GPU parallelization.


\section{Open Resources}
\label{sec:resources}

A central goal of HARL-A is to function as a community resource rather than a one-off implementation. This section describes the publicly available artifacts that support that goal.

\subsection{Pretrained Policies on Hugging Face}

Table~\ref{tab:policies} lists the pretrained policies released alongside this framework. All policies are hosted on Hugging Face and can be loaded directly via the HARL-A \gls{api} with a single configuration flag, without requiring users to retrain from scratch. Policies span curriculum stages (allowing inspection of how skills develop), team compositions (homogeneous and heterogeneous), and critic architectures (team-specific and shared-critic variants, the latter included explicitly to support ablation studies on the importance of critic isolation).

\begin{table}[h]
    \centering
    \footnotesize
    \begin{threeparttable}
    \begin{tabular}{lcc}
        \hline
        \textbf{Environment} & \textbf{Starting} & \textbf{Trained} \\
        \hline
        Leatherback-Stage1-Soccer-v0             & NO  & YES \\
        Leatherback-Stage2-Soccer-v0             & YES & YES \\
        AnymalC\_Soccer\_Hetero\_By\_Team-v0     & YES & NO  \\
        Sumo-Stage2-Hetero-By-Team-v0            & YES & NO  \\
        Sumo-Stage2-Hetero-Same-Critic-v0        & YES & NO  \\
        Sumo-Stage2-Hetero-Same-Critic-No-Neg-v0 & YES & NO  \\
        Sumo-Stage2-Hetero-v0                    & YES & YES \\
        Minitank-Adversarial-Direct-v0           & YES & YES \\
        Anymal-C-Go-To-Point-Sumo                & NO  & YES \\
        Isaac-Multi-Agent-Flat-AnymalC-Direct-v0 & YES & YES \\
        Anymal-C-Sumo-Stage1-Blocks-Push-v0      & YES & YES \\
        AnymalC\_Soccer\_Go\_To\_Point\_Stage\_0 & NO  & YES \\
        AnymalC\_Soccer\_Go\_To\_Ball\_Stage\_1  & YES & YES \\
        AnymalC\_Soccer\_Score\_Goals\_Stage\_2  & YES & YES \\
        leatherback-Sumo-Direct-MA-Stage1-v0     & NO  & YES \\
        AnymalC-VS-Leatherback-Soccer-v0         & YES & NO  \\
        \hline
    \end{tabular}
    \begin{tablenotes}
        \small
        \item ``Starting'' indicates a pretrained initialization checkpoint is provided; ``Trained'' indicates a fully trained policy is available.
    \end{tablenotes}
    \end{threeparttable}
    \caption{Pretrained policies available on Hugging Face. Environments span all three benchmark task categories, multiple curriculum stages, and both homogeneous and heterogeneous team configurations.}
    \label{tab:policies}
\end{table}

These policies serve several practical purposes. Researchers can use trained policies as fixed opponents for bootstrapping new agents, dramatically reducing the compute cost of initial training. Starting checkpoints enable fine-tuning on new morphologies or reward structures. The inclusion of shared-critic variants (e.g., \texttt{Sumo-Stage2-Hetero-Same-Critic-v0}) alongside team-specific variants provides a ready-made basis for ablation studies on the architectural choices described in Section~\ref{sec:framework}, without requiring users to reproduce the full training run.

\subsection{Codebase and Maintenance}

HARL-A is built as an extension of the HARL Python package \cite{zhong2024HARL}, which provides base implementations of heterogeneous \gls{marl} algorithms, including \gls{happo}, HATRPO, and supporting utilities. Our extension adds the adversarial training regime, team-indexed environment interface, and IsaacLab integration. Because HARL-A is structured as an additive layer on top of the HARL library rather than a fork, improvements to the underlying algorithm implementations are inherited automatically, and new algorithms from the HARL ecosystem can be tested against HARL-A environments with minimal integration work.

The codebase is open-source, actively maintained, and includes documentation for adding new environments, integrating new algorithms, and loading pretrained policies. The project page at https://directlab.dev/IsaacLab-HARL provides installation instructions, tutorials, and links to all released artifacts.

\section{Extending the Framework}
\label{sec:extension}

A core design goal of HARL-A is to impose as few constraints as possible on team structure, morphology, and algorithmic choice. Here we describe the extensibility concretely across four dimensions.

\noindent\textbf{Any number of teams.} The team-indexed \gls{api} places no upper bound on the number of competing teams. The three-team configuration shown in Figure~\ref{fig:multiple_environments} was implemented without any changes to the framework core---only the environment configuration file required modification. $N$-team competition, free-for-all scenarios, and asymmetric team sizes (e.g., 2-v-1-v-3) are all supported. It is important to note, as highlighted in figure~\ref{fig:fps}, that the speed of the training is heavily dependent on the number of articulations in the scene. 

\noindent\textbf{Any morphology mix.} IsaacLab's articulation system handles physics simulation for any robot described by a USD asset file. To add a new robot to an existing environment, a user specifies the asset path and defines the observation and action dimensions in the team configuration. The framework automatically constructs the corresponding actor network with the correct input/output dimensions, and the team-specific critic is sized accordingly. The 16 environments in Table~\ref{tab:policies} span quadrupeds, wheeled rovers, humanoids, drones, and tanks---all trained within the same framework without modifications to the core codebase.

\noindent\textbf{New environments with minimal boilerplate.} Isaaclab provides most of the scaffolding for implementing new environments and we provide our environments as open source resources for further adversarial environment development. 

\noindent\textbf{Algorithm integration.} The current release ships with \gls{happo} as the primary training algorithm. The team-indexed critic interface is compatible with any \gls{ctde} algorithm that accepts per-team value functions, meaning other algorithms from the HARL library (HATRPO, HAATRPO) and external implementations (value-decomposition \cite{son2019qtran}, graph-attention critics \cite{wei2024graphattentionnetworks}) can be tested against HARL-A benchmark environments by implementing a thin adapter layer.


\section{Limitations and Future Work}
\label{sec:limitations}

HARL-A's current release has several limitations that represent directions for future development. The framework currently integrates only \gls{happo}; broader algorithm coverage---including off-policy methods \cite{lerrel2017adversarialreinforcementlearning} and value-decomposition approaches \cite{son2019qtran}---would make it more useful as an algorithm benchmark. Evaluation methodology is currently limited to win rate against a fixed baseline; richer metrics such as exploitability, cross-play performance, and robustness to novel opponents would provide a more comprehensive picture of policy quality. The Soccer and 3D Galaga environments are at earlier stages of development than Sumo, and their benchmark status will strengthen with additional trained variants and multi-algorithm comparisons. Finally, the framework currently operates in simulation only; bridging to sim-to-real transfer for competitive tasks is an open challenge with significant practical importance.


\section{Conclusion}
\label{sec:conclusion}

We presented HARL-A, an open-source, actively maintained framework for scalable heterogeneous adversarial reinforcement learning in IsaacLab. By extending the HARL algorithm library with team-isolated critics and a team-indexed environment \gls{api}, HARL-A supports competitive training across any number of teams with any mix of robot morphologies---with no changes to the framework core required as team structure or morphology changes. The accompanying benchmark suite---Sumo, Soccer, and 3D Galaga---provides standardized environments spanning contact-rich pushing, ball-skill competition, and pursuit/evasion, and over ten pretrained policies on Hugging Face allow researchers to immediately explore adversarial dynamics, run cross-play evaluations, and use trained agents as opponents without retraining from scratch.

Our experiments confirm that the framework reliably produces learned adversarial policies across all three task types, and that morphological asymmetry alone induces emergent role specialization without any explicit role mechanism. GPU parallelization yields throughput sufficient to train competitive policies at $10^{10}$-step scale on commodity hardware.

We hope HARL-A serves as both a practical starting point and a community benchmark for adversarial \gls{marl} research in physically grounded settings, and that the combination of flexible infrastructure, open pretrained policies, and active maintenance lowers the barrier to entry for the broader robotics and multi-agent learning community.


\bibliography{references}
\bibliographystyle{ieeetr}

\end{document}